\documentclass[runningheads]{llncs}
\usepackage{graphicx}
\usepackage{epsfig}
\usepackage{dsfont}
\usepackage{amsmath}
\usepackage{amssymb}
\usepackage{boldline,multirow}
\usepackage{siunitx}
\usepackage{booktabs}
\usepackage{soul}
\usepackage{cite}
\usepackage{cases}
\usepackage{hyperref}
\usepackage{bbding}
\newcommand{\ucite}[1]{\cite{#1}} 
\begin{document}
\title{Conditional Training with Bounding Map for Universal Lesion Detection}
\titlerunning{Conditional Training with Bounding Map for Universal Lesion Detection}

%

\author{Han Li\inst{1,2} \and Long Chen\inst{2,3} \and Hu Han\inst{2}\Envelope \and S. Kevin Zhou\inst{2}\Envelope}
\institute{School of AI, University  of  the  Chinese  Academy  of  Science\\Medical Imaging, Robotics, Analytic Computing Laboratory/Engineering
(MIRACLE), Key Laboratory of Intelligent Information Processing of Chinese
Academy of Sciences (CAS), Institute of Computing Technology, CAS, Beijing, China\\ School of Electronic, Electrical and Communication Engineering, University  of  the  Chinese  Academy of  Science\\
\email{\{han.li,long.chen\}@miracle.ict.ac.cn, \{hanhu,zhoushaohua\}@ict.ac.cn}}
\authorrunning{H.~Li, H.~Han, and S.~Kevin Zhou}

\newcommand*\samethanks[1][\value{footnote}]{\footnotemark[#1]}


\authorrunning{Han~Li et al.}
\maketitle              
\begin{abstract}
{Universal Lesion Detection (ULD) in computed tomography plays an essential role in computer-aided diagnosis.  Promising ULD results have been reported by coarse-to-fine two-stage detection approaches, but such two-stage ULD methods still suffer from issues like imbalance of positive v.s. negative anchors during object proposal and insufficient supervision problem during localization regression and classification of the region of interest (RoI) proposals. While leveraging pseudo segmentation masks such as bounding map (BM) can reduce the above issues to some degree, it is still an open problem to effectively handle the diverse lesion shapes and sizes in ULD.
In this paper we propose a BM-based conditional training for two-stage ULD, which can (i) reduce positive vs. negative anchor imbalance via a BM-based conditioning (BMC) mechanism for anchor sampling instead of traditional IoU-based rule; and (ii) adaptively compute size-adaptive BM (ABM) from lesion bounding-box, which is used for improving lesion localization accuracy via ABM-supervised segmentation. Experiments with four state-of-the-art methods show that the proposed approach can bring an almost free detection accuracy improvement without requiring expensive lesion mask annotations.}

\end{abstract}

\section{Introduction}\label{sec:introduction}
Universal Lesion Detection (ULD)  in computed tomography (CT) \ucite{zlocha2019one-stage,tao2019improving,zhang2019anchor_free,zhang2020Agg_Fas,tang2019uldor,yan20183DCE,li2019mvp,yan2019mulan,yang2020alignshift,cai2020deep,li2020bounding,zhang2020revisiting,yan2020learning}, which aims to localize different types of lesions instead of identifying lesion types \ucite{liao2019evaluate,lin2019automated,wang2019volumetric,astaraki2019normal,tang2019nodulenet,shao2019attentive,liu20193dfpn,boot2020diagnostic,yu2020deep,fan2020positive,yu2020computer,de2020multi}, plays an essential role in computer-aided  diagnosis (CAD). ULD is a challenging task because different lesions have very diverse shapes and sizes, easily leading to false positive and false negative detections. Most existing ULD methods are mainly inspired by the successful deep models in object detection from natural images. These ULD approaches adapt the Mask-RCNN \cite{he2017maskrcnn} framework by constructing a pseudo segmentation mask for lesion regions as the extra supervision\ucite{zlocha2019one-stage,tao2019improving,tang2019uldor,yan2019mulan,yang2020alignshift,yan2020learning} or extract more 3D context information from multi CT slice\ucite{zhang2019anchor_free,zhang2020Agg_Fas,yan20183DCE,li2019mvp,yan2019mulan,yang2020alignshift,cai2020deep,zhang2020revisiting} to assist the ULD training.

\begin{figure}[t]
\centering
\setlength{\abovecaptionskip}{0.cm}

\includegraphics[scale=0.5]{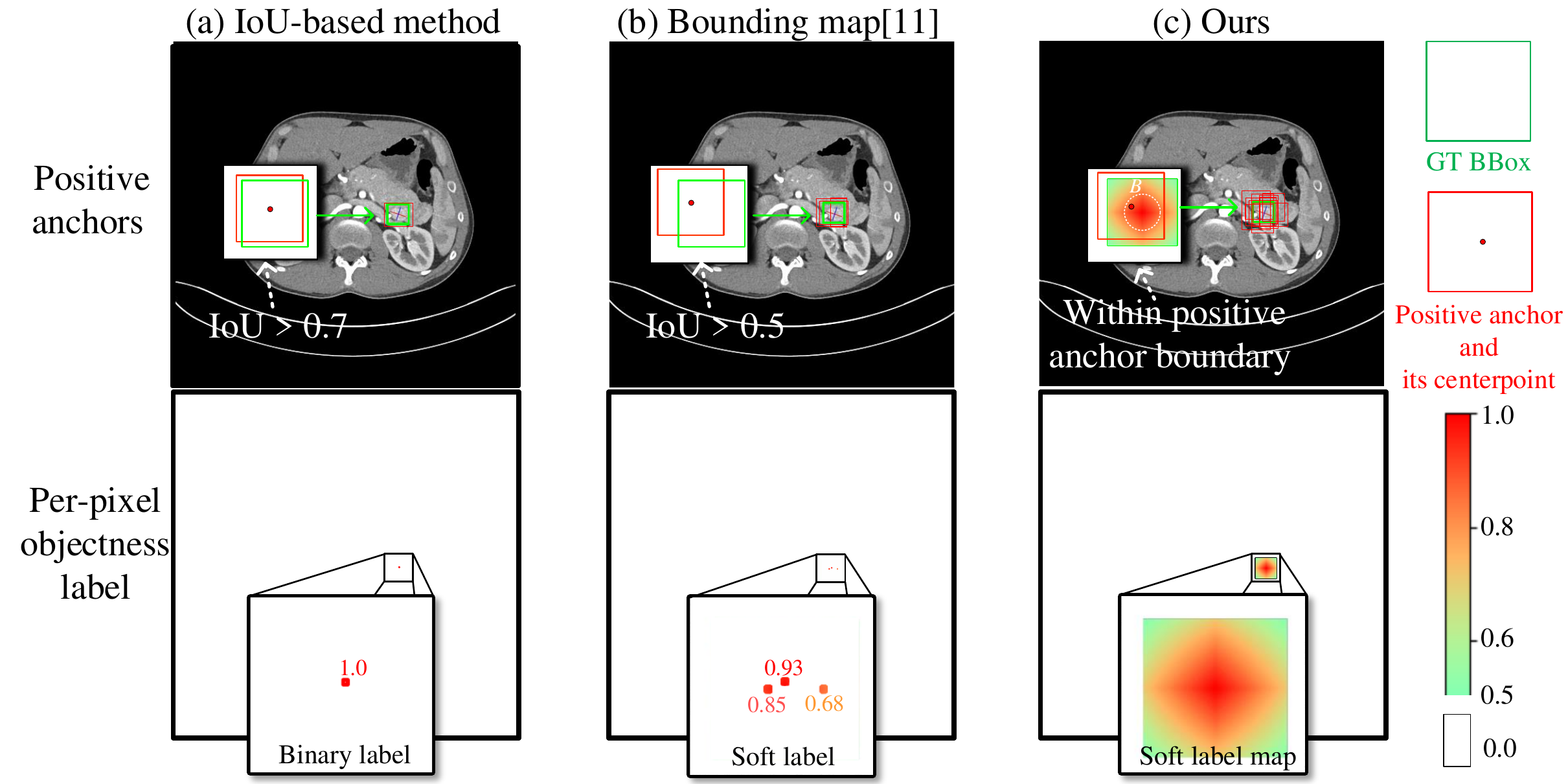}
\caption{The IoU-based method (a) produces very few positive anchors thus biases the RPN training. The BM method \cite{li2020bounding} (b) relieves this issue to some degree via a soft-label map BM and a lower threshold but the imbalance issue remains severe. Our BMC mechanism (c) predicts a per-pixel objectness map and any anchor within the BM with a value larger than a threshold can be a positive anchor. Thus, the positive vs. negative anchor imbalance issue can be effectively relieved.}
\label{fig:fig1_overview}
\end{figure}
Most of the above approaches proposed \ucite{tao2019improving,zhang2020Agg_Fas,tang2019uldor,yan20183DCE,li2019mvp,yan2019mulan,yang2020alignshift,zhang2020revisiting,yan2020learning} for ULD are designed based on a two-stage, anchor-based detection framework, i.e., proposal generation followed by classification and regression as in Faster R-CNN \cite{ren2015fasterrcnn}. 
While achieving good success, such a framework has inherent limitations: 
(i) \emph{Anchor imbalance in stage-1 \cite{oksuz2020imbalance}.}
In the first stage, anchor-based methods first find out the positive (lesion) anchors and use them as the region of interest (RoI) proposals according to the IoU between anchors and ground-truth (GT) BBoxs. An anchor is considered positive if its IoU with any GT BBox is greater than the IoU threshold and negative otherwise.  This idea helps natural images to get enough positive anchors because they may have a lot of GT BBoxs per image \cite{oksuz2020imbalance}, but it isn't suitable for ULD. Most CT slices only have one or two GT lesion BBox(s), so the amount of positive anchors of lesions is rather limited. This limitation can cause severe RoI imbalance (our empirical statistics show the positive vs. negative proposal imbalance can be as large as 1:200 which is showed in Suppl. Material), thereby leading to difficulty in network convergence.
(ii) \emph{Insufficient supervision in stage-2.} 
In the second stage, each RoI proposal from the first stage gets a classification score indicating its probability of lesion vs. non-lesion. The RoI proposals with high classification scores are chosen to obtain the final lesion BBox predictions. Since various lesions may have similar appearances with the other tissues; the  RoI proposals from non-lesion regions can also get very high lesion classification scores. Hence, a single classification score cannot provide sufficient supervision information to remove the false detections in the second stage. 

To tackle the above two challenges, recently Li et al. \cite{li2020bounding} propose a bounding map (BM) based two-stage ULD method, which uses a lower IoU threshold to determine anchor categories based on IoU mechanism, and uses the value of BM as the soft labels. In addition, the BMs are also used as pseudo lesion masks for a newly introduced segmentation branch in stage-2. However, such an approach still suffers from several issues: (i) generating BMs from BBox in a fixed manner without considering lesion size and shape differences leads to sub-optimum BM representation for small lesions; and (ii) using IoU to determine anchor categories may lead to incorrect classification for lesions with irregular shapes.


To address the above issues, while exploiting the advantages of the BM-based two-stage ULD method, we propose a novel training mechanism for ULD to effectively reduce positive vs. negative anchor imbalance via a BM-based conditioning (BMC) mechanism in stage-1 and improve lesion localization accuracy by leveraging a size-adaptive BM (ABM) for supervising the segmentation branch in stage-2.
Different from the traditional IoU-based positive and negative anchors selection mechanism, we use two independent anchor classification and regression tasks by directly predicting a \textbf{per-pixel objectness map} and selecting positive anchors for regression based on $BM$s \cite{li2020bounding}. Specifically as shown in Fig. \ref{fig:fig1_overview}, we use the $BM_{xy}$ \cite{li2020bounding} as objectness GT map and select anchors whose value is greater than a threshold $B$ in $BM_{xy}$ to the region proposal network (RPN) for pixel-wise objectness map regression. We further randomly mask out some background pixels in $BM_{xy}$ during objectness training to keep the number of background pixels no more than two times of the foreground pixel number.
In addition, we extend BM into size-adaptive BMs (ABMs) to handle diverse lesions, in which the small lesions will be enhanced to compensate its limited pixel number while the big lesions are weakened. We use an ABM-supervised segmentation branch in stage-2 to improve the lesion localization accuracy.

Our method works perfectly with two-stage ULD methods, so it can be integrated with the state-of-the-art (SOTA) two-stage ULD methods. We conduct extensive experiments on the DeepLesion dataset \cite{yan18deeplesion} with three SOTA ULD methods to validate the effectiveness of our method.
\begin{figure}[t]
\centering
\setlength{\abovecaptionskip}{0.cm}

\includegraphics[scale=0.33]{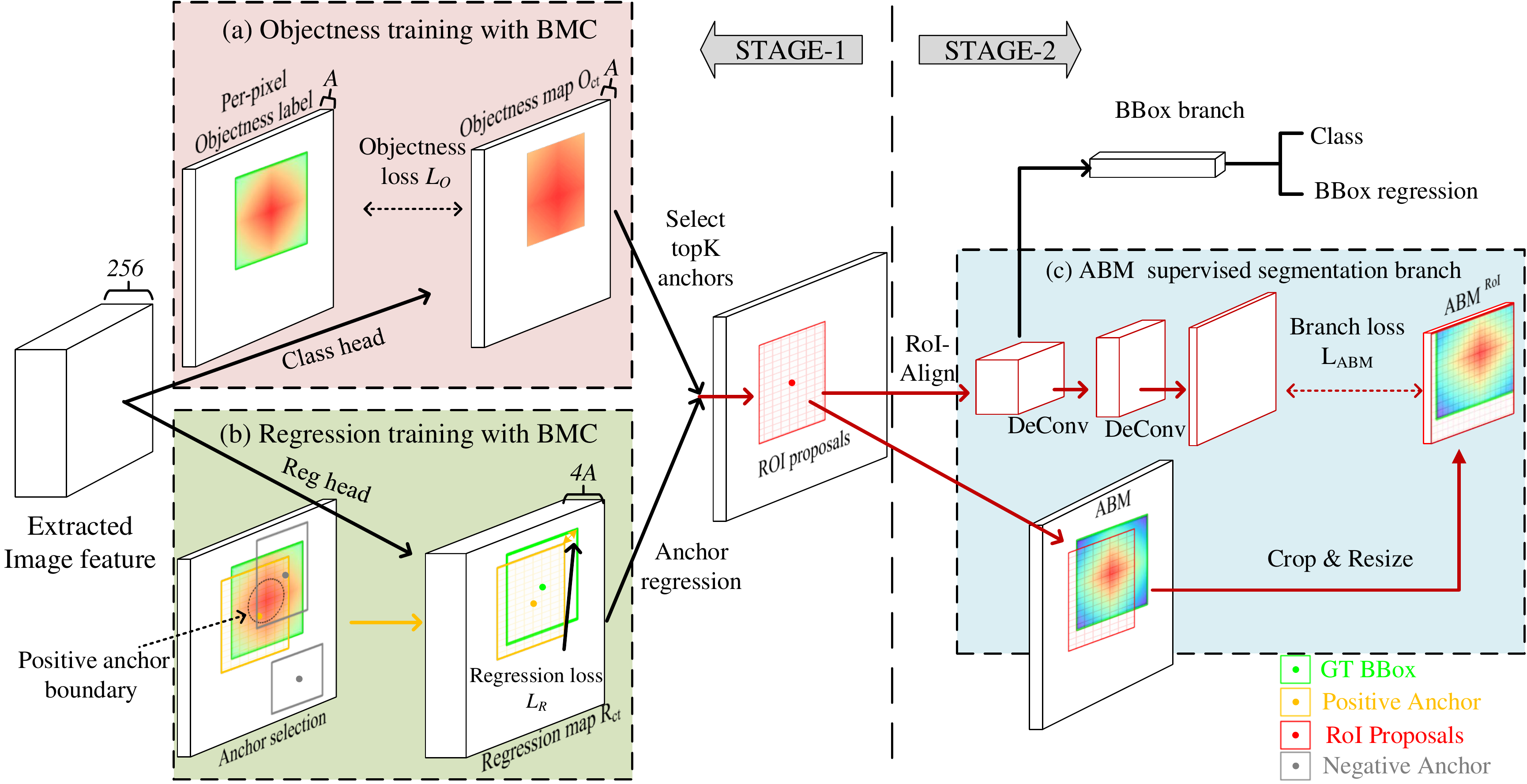}
\caption{The network architecture of the proposed approach, in which, BMC mechanism (a\&b) and ABM branch (c) are respectively used in the two stages of the ULD method.}

\label{fig:fig2_network_architecture}
\end{figure}

\section{Method}
As shown in Fig. \ref{fig:fig2_network_architecture}, we utilize the BMC mechanism ((a) \& (b) in Fig. \ref{fig:fig2_network_architecture}) to reduce positive vs. negative anchor imbalance  in stage-1 and improve lesion localization accuracy by adding a supervised segmentation branch based on size-adaptive BM (ABM) ((c) in Fig. \ref{fig:fig2_network_architecture}) in stage-2. Section \ref{sec:bm} details the BM generation process \cite{li2020bounding}; Section \ref{sec:bmc} introduces the BMC mechanism, and Section \ref{sec:abm} explains the newly introduced ABM-supervised segmentation branch.

\subsection{Bounding map generation}\label{sec:bm}
For the $n^{th}$ GT BBox in an image $I_{ct}$, the BMs $BM^{(n)}_x$ and $BM^{(n)}_y$ are computed based on an all-zero map, in which the element values within the BBox at the same location as the GT BBox are assigned to values linearly interpolated from 1 to 0.5. The all-zero map and the final BMs have the same size as the image feature map output by the backbone network.  Specifically,
\begin{equation}
BM^{(n)}_x(x,y)=\left\{
\begin{array}{lp{8mm}<{\centering}l}
0& & {(x,y) \notin S^{(n)}_{BBox}}\\
1-k^{(n)}_x\left|x^{(n)}-x^{(n)}_{ctr}\right| & &{(x,y) \in S^{(n)}_{BBox}}
\end{array} \right. ,
\end{equation}

\begin{equation}
BM^{(n)}_y(x,y)=\left\{
\begin{array}{lp{8mm}<{\centering}l}
0& & {(x,y) \notin S^{(n)}_{BBox}}\\
1-k^{(n)}_y\left|y^{(n)}-y^{(n)}_{ctr}\right| & &{(x,y) \in S^{(n)}_{BBox}}
\end{array} \right. ,
\end{equation}
where $S^{(n)}_{BBox}$ denotes the $n^{th}$ GT BBox $(x^{(n)}_1,y^{(n)}_1,x^{(n)}_2,y^{(n)}_2)$, whose center is at $(x^{(n)}_{ctr},y^{(n)}_{ctr})$. The slope $k^{(n)}$ calculation is defined as:
\begin{equation}
  k^{(n)}_x= \frac{1}{ x^{(n)}_2-x^{(n)}_{1}}, ~~k^{(n)}_y= \frac{1}{ y^{(n)}_2-y^{(n)}_{1}},
\end{equation}
and the final $BM_x$ ($BM_y$) is generated by aggregating all the $BM^{(n)}_x$s ($BM^{(n)}_y$s):

 \begin{equation}
   BM_x=\min \Big [ \sum\limits_{i=1}^I BM^{(n)}_x,1 \Big ], BM_y=\min  \Big [\sum\limits_{i=1}^I BM^{(n)}_y, 1  \Big ],
  \end{equation}
where $N$ is the number of GT BBoxs of a CT image. Finally, the $BM_{xy}$ is  formed by  element-wise multiplication between $BM_x$ and $BM_y$\cite{li2020bounding}:
\begin{equation}
  BM_{xy}=\sqrt[2]{BM_{x}\odot BM_{y} },
\end{equation}
Different from \cite{li2020bounding}, we use only $BM_{xy}$ and we will call it $BM$ below.
\subsection{BM-based conditioning mechanism} \label{sec:bmc}
The RPN is trained to produce object bounding map $R_{ct} \in \mathcal{R}^ {\frac{W}{R}\times \frac{H}{R} \times 4A}$ via regression and objectness score map $O_{ct} \in \mathcal{R}^ {\frac{W}{R}\times \frac{H}{R} \times A}$ via classification at each position (or an anchor's centerpoint) in an input image $I_{ct}\in \mathcal{R}^ {W \times H}$, where $R$ and $A$ are the network output stride and the number of anchor classes, respectively. In previous ULDs, anchors are selected for training objectness and regression in conventional RPN; here we develop two independent BM-based processes for RoI proposal: (i) objectness map prediction with BMC and (ii) BBox regression with BMC.

\textbf{Objectness map prediction with BMC.}
Conventionally, the objectness GT labels of positive, negative and otherwise anchors are set as 1, 0, and -1, respectively, and only the positive and negative anchors are used for loss calculation.
Motivated by FCOS \cite{zhou2019objectsaspotints}, we first resize $BM\in \mathcal{R}^ {W\times H \times A}$ into $BM^{r} \in \mathcal{R}^ {\frac{W}{R}\times \frac{H}{R} \times A}$ in a linear interpolation manner, then directly use $BM^{r}$ as the objectness GT lable and train RPN to predict the objectness map in a per-pixel manner as showed in Fig. \ref{fig:fig2_network_architecture} (a).
Considering the lesion sizes variation and foreground-background  pixel imbalance, we further randomly ignore some background pixels in $BM^{r}$ to keep the number of background pixels no more than two times of the foreground pixel number $N_{f}$. Specifically, we first find out the foreground and background pixels set $S_f$ and $S_b$ based on $BM^r$:
\begin{equation}
  S_f=\{(x_f,y_f)|BM^{r}_{xy}(x_f,y_f)\geq 0.5\},~S_b=\{(x_b,y_b)|BM^{r}_{xy}(x_b,y_b) \textless 0.5\}.
\end{equation}
Then the number of foreground pixels $N_{f}= Card(S_f)$ can be counted.
After that, we randomly sample $2 N_f$ background pixels from the background pixels set $S_b$ as the training set $S^t_b$. 
Finally, we train the RPN for predicting objectness map $O_{ct}\in \mathcal{R}^ {\frac{W}{R}\times \frac{H}{R} \times A}$ with a binary cross entropy (BCE) loss:
\begin{equation}
 \mathcal{L}_{o}=
\begin{array}{lp{2mm}<{\centering}l}
\sum\limits_{j=1}^J\mathcal{L}_{BCE}(O_{ct}(x_j,y_j),BM^{r}_{xy}(x_j,y_j))&&{(x_j,y_j) \in S^t_b\cup S_f}
\end{array} ,
\end{equation}
where $J$ is the total number of pixels in $S^t_b\cup S_f$.

\textbf{BBox regression with BMC.}
In our method, the loss computation  during BBox regression is still applied to positive anchors, but we propose for BBox regression a new positive anchor selection method based on BM (see Fig. \ref{fig:fig2_network_architecture} (b)). We first define a positive anchor boundary value $B$  and set the anchors whose corresponding $BM^{r}_{xy}$ value is greater  than $B$ as positive  anchors:
\begin{equation}\label{equ:equ8}
  Loc^{(n)}_p=\left\{
  \begin{array}{lp{8mm}<{\centering}l}
    \{x^{(n)}_p,y^{(n)}_p|BM^{r(i)}_{xy}(x^{(n)}_p,y^{(n)}_p)\geq B\} & &A^{(n)}_{BBox} \geq 16\\
\{x^{(n)}_p,y^{(n)}_p|BM^{r(i)}_{xy}(x^{(n)}_p,y^{(n)}_p)\geq 0.5\} & & else
\end{array} \right. ,
\end{equation}
where $Loc^{(n)}_p$ and $A^{(n)}_{BBox}$ is the positive anchor's centerpoint location set for the $n^{th}$ GT BBox and the area of the $n^{th}$ GT BBox. Finally, we union all $Loc^{(n)}_p$ and get $Loc_p$, the anchor centerpoint location set for the image.

Finally the BBox regression loss can be written as:
\begin{equation}
 \mathcal{L}_{r}=
\begin{array}{lp{2mm}<{\centering}l}
\sum\limits_{m=1}^M\mathcal{L}_{reg}(x_m,y_m)&&{(x_m,y_m) \in Loc_p}
\end{array} ,
\end{equation}
where $\mathcal{L}_{reg}$ is the original regression loss function and $M$ is the number of anchors' centerpoint locations in $Loc_p$.

\begin{figure}[t]
  \centering
  \setlength{\abovecaptionskip}{-0.05cm}

   \includegraphics[scale=0.5]{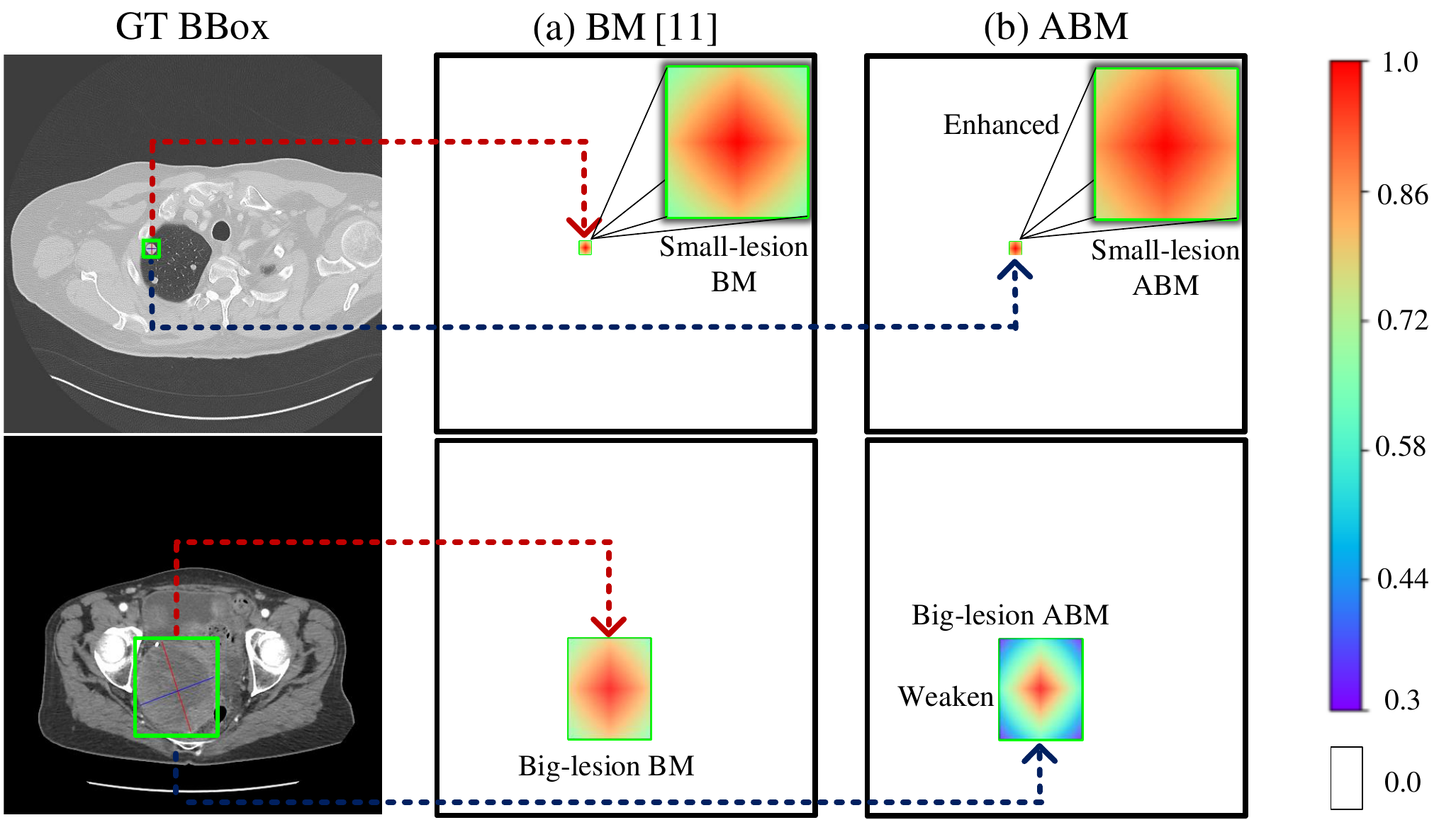}
  \caption{Different from the BM (a) that uses a fixed generation method \cite{li2020bounding}, the proposed ABM (b) can enhance the small lesions while weaken the big lesions.} 
  \label{fig:fig3_ABM}

\end{figure}
\subsection{Pseudo lesion segmentation via ABM supervision.} \label{sec:abm}
The top-K objectness anchors regressed by the regression are used as RoI proposals and the input of the second-stage classification and regression. However, as we analyzed in Sec. \ref{sec:introduction}, the BBox branch does not provide enough supervision for the complex ULD task. We introduce an extra branch, i.e., ABM supervised pseudo lesion segmentation branch.

\textbf{Size adaptive bounding maps (ABM).}
As shown in Section \ref{sec:bm}, BM uses a fixed generating manner without considering lesion size and shape differences leads to sub-optimum representations for small lesions. As shown in Fig. \ref{fig:fig3_ABM}, we extend BM to make it adaptive to lesion size, so that a small lesion is enhanced to compensate its limited area. Specifically, we multiply the two slopes $k^{(n)}_x$ and $k^{(n)}_y$ by a ratio $\alpha$ when generating the $ABM^{(n)}_x$ and $ABM^{(n)}_y$:
\begin{equation}
ABM^{(n)}_x(x,y)=\left\{
\begin{array}{lp{8mm}<{\centering}l}
0& & {(x,y) \notin S^{(n)}_{BBox}}\\
1-\alpha k^{(n)}_x\left|x^{(n)}-x^{(n)}_{ctr}\right| & &{(x,y) \in S^{(n)}_{BBox}}
\end{array} \right. ,
\end{equation}
\begin{equation}
ABM^{(n)}_y(x,y)=\left\{
\begin{array}{lp{8mm}<{\centering}l}
0& & {(x,y) \notin S^{(n)}_{BBox}}\\
1-\alpha k^{(n)}_y\left|y^{(n)}-y^{(n)}_{ctr}\right| & &{(x,y) \in S^{(n)}_{BBox}}
\end{array} \right. ,
\end{equation}
\begin{equation}\label{equ:equ12}
\alpha=\left\{
\begin{array}{lp{8mm}<{\centering}l}
0& & A^{(n)}_{BBox} \textless A_s\\
1& &  A_s \leq A^{(n)}_{BBox}\textless A_m \\
1.4 & &A^{(n)}_{BBox}\geq A_m
\end{array} \right. ,
\end{equation}
where the $A_s, A_m$ are the size thresholds for small and medium lesions.
The final $ABM_{xy}$ generation is the same as the $BM_{xy}$.

\textbf{Segmentation with ABM supervision.}
As shown in Fig. \ref{fig:fig2_network_architecture} (c), the ABM  supervised segmentation branch is the same as the mask branch in Mask R-CNN \cite{he2017maskrcnn}. It is parallel to the BBox classification and regression branch. The ABM is first cropped based on the RoI BBox and resized to the size of the output to obtain $ABM^{RoI}_x \in \mathcal{R}^ {W_b\times H_b\times 1}$, where $W_b$ and $H_b$ are the output  $\hat{ABM}^{RoI}$'s width and height of ABM-supervised segmentation branch, respectively. Then the loss function of ABM branch for each RoI can be defined as a norm-2 loss:
\begin{equation}
\mathcal{L}_{ABM}=\mathcal{L}_2(\hat{ABM}^{RoI},ABM^{RoI}).
\end{equation}

\section{Experiments}
\subsection{Dataset and setting}
We conduct experiments on the DeepLesion dataset \cite{yan18deeplesion}.  The dataset contains 32,735 lesions on 32,120 axial slices from 10,594 CT studies of 4,427 unique patients. Most existing datasets typically focus on one type of lesion, while DeepLesion contains a variety of lesions with large diameters ranges (from 0.21 to 342.5mm). The 12-bit intensity CT is rescaled to [0,255] with different window ranges settings used in different frameworks. Also, every CT slice is resized and interpolated according to the detection frameworks' setting. We follow the official split, i.e., $70\%$ for training, $15\%$ for validation and $15\%$ for testing. The number of false positives per image (FPPI) is used as the evaluation metric. We set the positive anchor boundary value $B$ in Equ. \ref{equ:equ8} as 0.25, size thresholds $S_s$ and $S_m$ in Equ. \ref{equ:equ12} as 250 and 1000 respectively. For training, we use the original network architecture and settings, and initialize the network using pretrained ULD models on DeepLesion dataset \cite{yan18deeplesion}.
\newsavebox{\tablebox}
\subsection{Lesion detection performance}

We apply our method with four SOTA ULD approaches \ucite{yan20183DCE,li2019mvp,yan2019mulan,yang2020alignshift} to evaluate the effectiveness. We also compare with three SOTA anchor-free  \ucite{tian2019fcos,zhou2019objectsaspotints,zhu2020deformable}  and  one two-stage anchor-based  \ucite{ren2015fasterrcnn} detection methods. As shown in Table \ref{results},  our method brings promising detection performance improvements for all baselines. The improvements of Faster R-CNN \cite{ren2015fasterrcnn}, 9-slice 3DCE, and MVP-Net are more pronounced than those of MULAN w/o SRL\cite{yan2019mulan} and AlignShift \cite{yang2020alignshift}. This is because MULAN and AlignShift introduce \underline{extra weakly segmentation} mask generated from radiologist-annotated RECIST labels.  The anchor-free methods get unsatisfactory results mainly because they are lack of the initialize advantage of anchors and the coarse-to-fine training advantage of the two-stage mechanism. We also provide a case to show our method's effectiveness in Suppl. Material.

\begin{table}[t]
\centering
\caption{Sensitivity (\%) at various FPPI on the testing set of DeepLesion \cite{yan18deeplesion}.} \label{results}
\begin{lrbox}{\tablebox}
\begin{tabular}{p{58mm}p{8mm}<{\centering}p{22mm}<{\centering}p{23mm}<{\centering}p{23mm}<{\centering}p{23mm}<{\centering}p{23mm}<{\centering}}
\toprule[1.5pt]
\textbf{Methods}&\textbf{slices}&\textbf{$@0.5$}&\textbf{$@1$}&\textbf{$@2$}&\textbf{$@4$}&Avg.[0.5,1,2,4]\\
\toprule[1pt]
Faster R-CNN \cite{ren2015fasterrcnn}&3&57.17 & 68.82 &74.97 &82.43&70.85\\
Faster R-CNN+BM\cite{li2020bounding} &3&63.96 (6.79$\uparrow$) & 74.43 (5.61$\uparrow$) &79.80 (4.83$\uparrow$) &86.28  (3.85$\uparrow$)&76.12(5.27$\uparrow$) \\
Faster R-CNN+Ours &3&65.37 (\textbf{8.20}$\uparrow$) & 76.31 (\textbf{7.49}$\uparrow$) &81.03 (\textbf{6.06}$\uparrow$) &87.98  (\textbf{5.55}$\uparrow$)&77.67(\textbf{6.82}$\uparrow$) \\
\midrule
3DCE  \cite{yan20183DCE} &9 &59.32 &70.68&79.09&84.34&73.36\\
3DCE+BM\cite{li2020bounding} &9&64.38 (5.06$\uparrow$)&75.55 (4.87$\uparrow$) &82.74  (3.65$\uparrow$)& 87.78 (3.44$\uparrow$)&77.62(4.26$\uparrow$)\\
3DCE+Ours &9&66.98 (7.66$\uparrow$)&77.25 (6.57$\uparrow$) &83.64  (4.55$\uparrow$)& 88.41 (4.07$\uparrow$)&79.07(5.71$\uparrow$)\\
\midrule
MVP-Net  \cite{li2019mvp}&9&70.07&78.77&84.91&87.33&80.27\\
MVP-Net+BM\cite{li2020bounding}&9&72.12 (2.05$\uparrow$)&80.51 (1.74$\uparrow$) &86.08 (1.17$\uparrow$)& 88.41 (1.08$\uparrow$)&81.78(1.51$\uparrow$)\\
MVP-Net+Ours&9&73.05 (2.98$\uparrow$)&81.41 (2.64$\uparrow$) &87.22 (2.31$\uparrow$)& 89.37 (2.04$\uparrow$)&82.76(2.49$\uparrow$)\\
\midrule
MULAN w/o (SRL\&RECIST) \cite{yan2019mulan}&9&72.34&80.17&85.21&89.41&81.78\\
MULAN w/o (SRL\&RECIST) +BM\cite{li2020bounding}&9&74.71 (2.37$\uparrow$)&81.62 (1.45$\uparrow$) &86.44 (1.23$\uparrow$)& 89.82 (0.41$\uparrow$)&83.15(1.37$\uparrow$)\\
MULAN w/o (SRL\&RECIST) +Ours&9&75.97 (3.63$\uparrow$)&82.81 (2.64$\uparrow$) &86.85 (1.64$\uparrow$)& 90.00(0.59$\uparrow$)&83.91(2.13$\uparrow$)\\
\midrule
MULAN w/o SRL   \cite{yan2019mulan}&9&73.85&81.02&85.98&90.01&82.71\\
MULAN w/o SRL +Ours&9&76.28 (2.43$\uparrow$)&83.13 (2.11$\uparrow$) &87.30 (1.32$\uparrow$)& 90.49 (0.48$\uparrow$)&84.30(1.59$\uparrow$)\\
\midrule
AlignShift w/o RECIST \cite{yang2020alignshift}&7&76.31&84.07&88.08&91.62&85.02\\
AlignShift w/o RECIST+BM\cite{li2020bounding}&7&77.51 (1.20$\uparrow$)&84.95 (0.88$\uparrow$) &88.63 (0.55$\uparrow$)& 92.08 (0.46$\uparrow$)&85.79(0.77$\uparrow$)\\
AlignShift w/o RECIST+Ours&7&79.03 (2.72$\uparrow$)&85.63 (1.56$\uparrow$) &89.53 (1.45$\uparrow$)& 92.43 (0.81$\uparrow$)&86.66(1.64$\uparrow$)\\
\midrule
AlignShift \cite{yang2020alignshift}&7&77.20&84.38&89.03&92.31&85.73\\
AlignShift +Ours&7&\textbf{79.17} (1.97$\uparrow$)&\textbf{85.71} (1.33$\uparrow$) &\textbf{89.80} (0.77$\uparrow$)& \textbf{92.65} (0.34$\uparrow$)&\textbf{86.83}(1.10$\uparrow$)\\
\toprule[1pt]
FCOS (anchor-free) \cite{tian2019fcos}&3&37.78&54.84&64.12&77.84&58.65\\
Objects as points (anchor-free)\cite{zhou2019objectsaspotints}&3&34.87&43.58&52.41&64.01&48.72\\
Deformable detr (anchor-free) \cite{zhu2020deformable}&3&57.62&65.64&70.65&75.58&67.37\\
\bottomrule[1.5pt]
\end{tabular}

\end{lrbox}
\scalebox{0.65}{\usebox{\tablebox}}
\end{table}

\subsection{Ablation study}
Ablation study is provided to evaluate the importance of the three key components of the proposed method: (i) Objectness map prediction with BMC, (ii) Box regression with BMC, and (iii) $ABM$ supervised segmentation.  As shown in Table \ref{ablation_study},  using objectness map prediction with BMC, we obtain a 1.79\% improvement over the baseline. Further adding BBox regression training with BMC accounts for another 0.18\% improvement and gives the best performance. Without using \underline{extra RECIST label}, $ABM$ and $BM$ increases the performance by 1.08\% and 0.02\%, respectively. In addition, the use of our method brings a minor increase (less than 10\%) in the inference time on a Titan RTX GPU.

\begin{table}[t]
\centering
\caption{Ablation study of our method at various FPs per image (FPPI).} 
\label{ablation_study}
\begin{lrbox}{\tablebox}

\begin{tabular}{p{20mm}<{\centering}p{25mm}<{\centering}p{25mm}<{\centering}p{25mm}<{\centering}p{18
mm}<{\centering}p{18mm}<{\centering}|p{18mm}<{\centering}p{18mm}<{\centering}|p{25mm}<{\centering}}
\toprule
AlignShift \cite{yang2020alignshift}&RECIST label&BMC objectness training &BMC regression training&ABM branch&BM branch \cite{li2020bounding} &$FPPI$=0.5&$FPPI$=1&Inference (s/img)\\
$\checkmark$&$\checkmark$&&&&&77.20&84.38&0.1758\\
$\checkmark$&$\checkmark$&$\checkmark$&&&&78.99&85.65&0.1747\\
$\checkmark$&$\checkmark$&$\checkmark$&$\checkmark$&&&\textbf{79.17}&\textbf{85.71}&0.1833\\
\midrule
$\checkmark$&&$\checkmark$&$\checkmark$&&&77.95&84.99&0.1784\\
$\checkmark$&&$\checkmark$&$\checkmark$&$\checkmark$&&\textbf{79.03}&\textbf{85.63}&0.1814\\
$\checkmark$&&$\checkmark$&$\checkmark$&&$\checkmark$&77.97&85.04&0.1755\\
\bottomrule
\end{tabular}

\end{lrbox}
\scalebox{0.5}{\usebox{\tablebox}}
\end{table}

\section{Conclusion}
In this paper, we try to overcome the two intrinsic limitations of two-stage ULD methods: anchor imbalance in stage-1 and insufficient supervision in stage-2. To relieve these, we propose a new BMC mechanism in stage-1 and an ABM supervised segmentation branch in stage-2.
Extensive experiments using several SOTA baselines on the DeepLesion dataset show that our approach can effectively boost the ULD performance with almost no additional computational cost.

%

%

\newpage
\bibliographystyle{splncs04}
\bibliography{egbib}

\end{document}


\title{Supplementary Materials for \\Conditional Training with Bounding Map for Universal Lesion Detection}
\titlerunning{~}

\author{Paper ID 487}
\institute{Anonymous Organization}
\authorrunning{Paper ID 487}

\maketitle              

\section{Visualization of detection results}
\begin{figure}[h]
\centering

\includegraphics[scale=0.32]{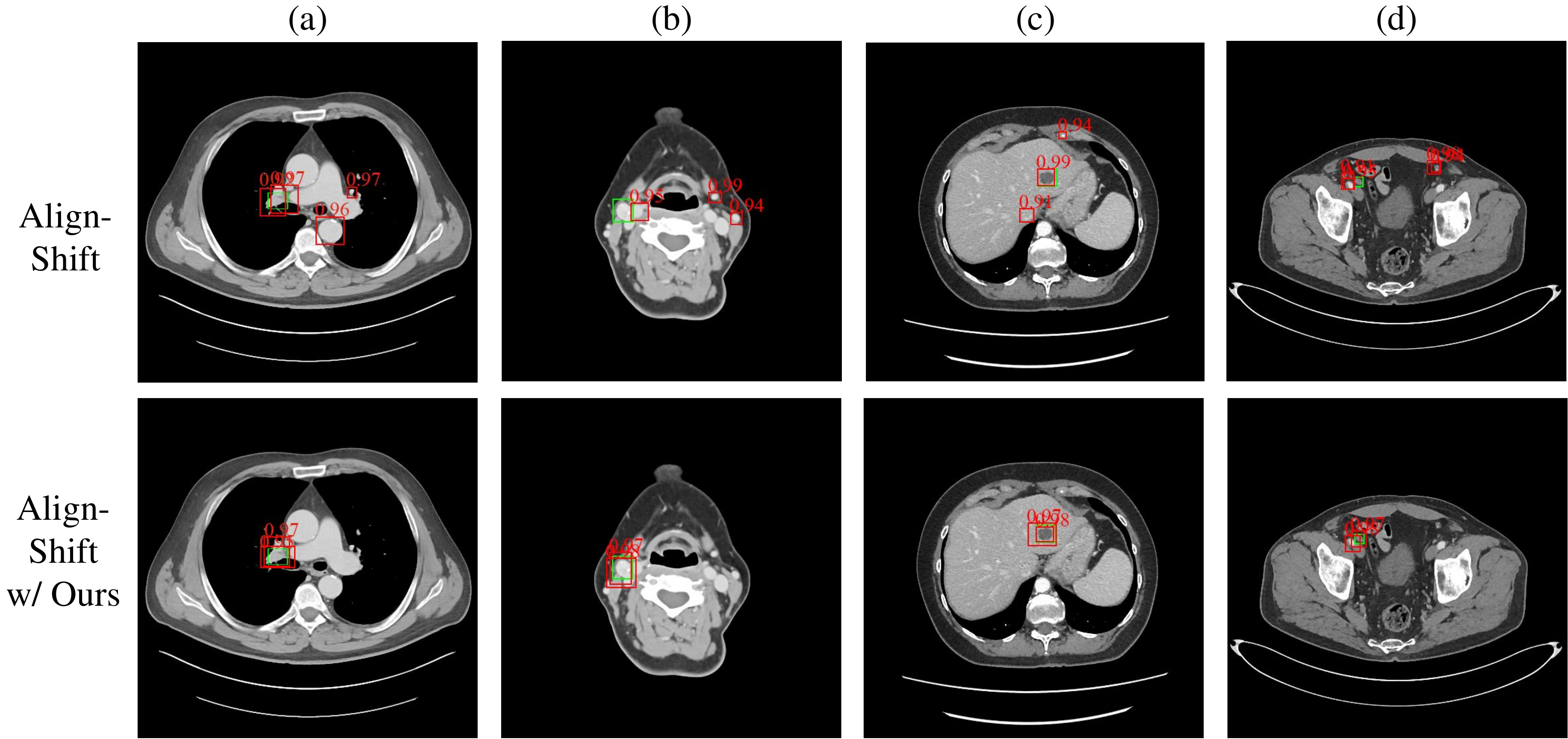}
\caption{High score BBoxs prediction (above 0.9) of AlignShift \cite{yang2020alignshift} with or without proposed method on test images. Red and green BBoxes denotes predicted BBoxs and GT BBoxs, respectively. The scores are marked above the BBoxs.}
\label{fig:fig4_vis_results}
\end{figure}

We randomly choose four CT images from DeepLesion dataset \cite{yan18deeplesion} to visually compare the detection results of AlignShift \cite{yang2020alignshift} w/ and w/o our methods. As shown in Fig. \ref{fig:fig4_vis_results}, the number of  high-classification-score FP predictions is decreased (a\&c\&d) and more accurate results (b) are obtained with the help of our method.

\begin{figure}[t]
\centering
\includegraphics[scale=0.4]{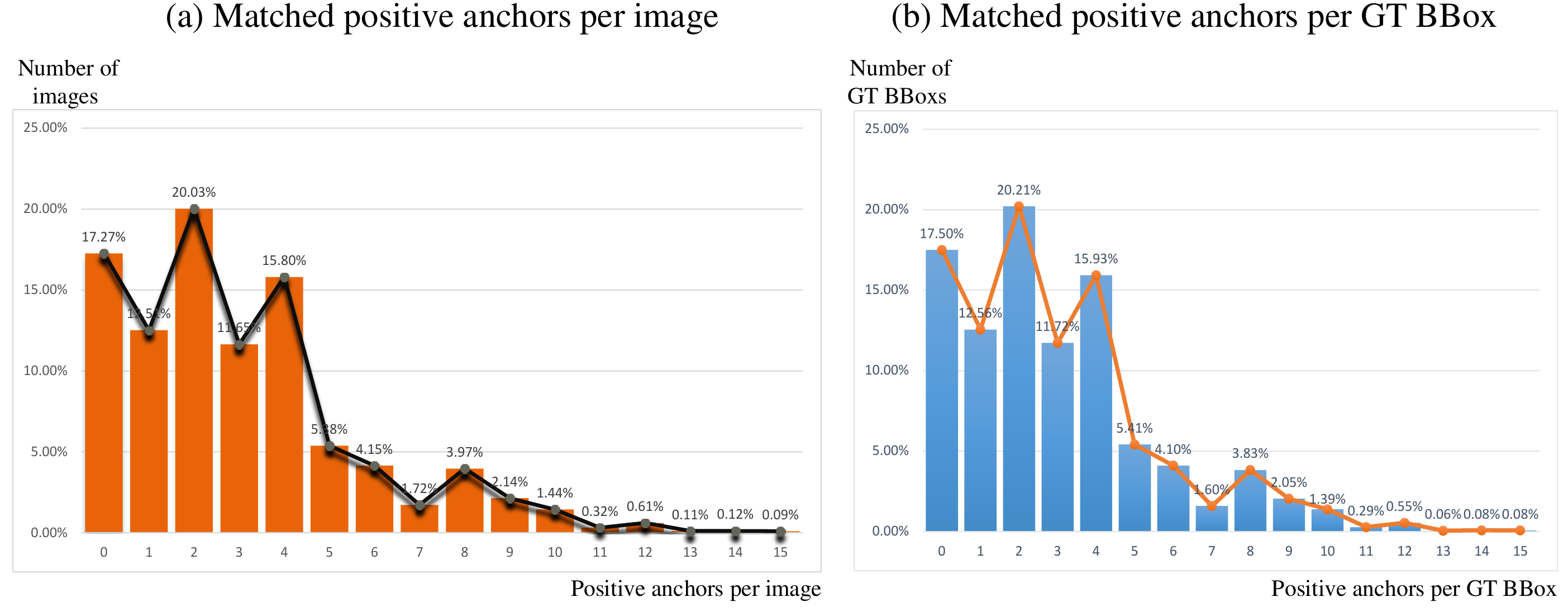}
\caption{We count the number of matched positive anchors per image (a) and per GT BBoxs (b) based on the IoU mechanism.}
\label{fig:fig4_anchors}
\end{figure}

\section{The number of positive anchors based on IoU}
We compute the number of matched positive anchors per image (a in Fig. \ref{fig:fig4_anchors}) and per GT BBoxs (b in Fig. \ref{fig:fig4_anchors}) based on the IoU mechanism. The IoU threshold is set as 0.7. As shown in Fig. \ref{fig:fig4_anchors}, the number of positive anchors is very few for both per-image manner and per-GT-BBox manner. More than 70$\%$ of GT-BBoxs (or images) match no greater than four positive anchors. This will cause a high imbalance of positive anchors v.s. negative anchors during training.
\bibliographystyle{splncs04}
\bibliography{egbib}